\documentclass[journal]{IEEEtran}
\usepackage{xcolor,soul,framed} 

\colorlet{shadecolor}{yellow}
\usepackage[pdftex]{graphicx}
\graphicspath{{../pdf/}{../jpeg/}}
\DeclareGraphicsExtensions{.pdf,.jpeg,.png}

\usepackage[cmex10]{amsmath}
\usepackage{array}
\usepackage{mdwmath}
\usepackage{mdwtab}
\usepackage{eqparbox}
\usepackage{url}

\begin{document}
\bstctlcite{IEEEexample:BSTcontrol}
    \title{An Improved ResNet50 Model for Predicting Pavement
Condition Index (PCI) Directly from Pavement Images}
  \author{Andrews~Danyo,~\IEEEmembership{Student Member,~IEEE,}
      Anthony~Dontoh,~\IEEEmembership{Member,~IEEE,}\\
      Armstrong~Aboah,~\IEEEmembership{Member,~IEEE,}

  \thanks{Manuscript received July 10, 2012. This paper is an expanded paper from the IEEE MTT-S Int. Microwave Symposium held on June 17-22, 2012 in Montreal, Canada. This work was funded in part by the Office of Naval Research under the Defense Advanced Research Projects Agency (DARPA) Microscale Power Conversion (MPC) Program under Grant N00014-11-1-0931, and in part by the Advanced Research Projects Agency-Energy (ARPA-E), U.S. Department of Energy, under Award Number DE-AR0000216.}
  \thanks{M. Roberg is with TriQuint Semiconductor, 500 West Renner Road Richardson, TX 75080 USA (e-mail: michael.roberg@tqs.com).}
  \thanks{T. Reveyrand is with the XLIM Laboratory, UMR 7252, University of Limoges, 87060 Limoges, France (e-mail: tibault.reveyrand@xlim.fr).}%
  \thanks{I. Ramos and Z. Popovic are with the Department of Electrical, Computer and Energy Engineering, University of Colorado, Boulder, CO, 80309-0425 USA (e-mail: ignacio.ramos@colorado.edu; zoya.popovic@colorado.edu).}
  \thanks{E. Falkenstein is with Qualcomm Inc., 6150 Lookout Road
Boulder, CO 80301 USA (e-mail: erez.falkenstein@gmail.com).}}

\markboth{
}{Roberg \MakeLowercase{\textit{et al.}}: High-Efficiency Diode and Transistor Rectifiers}

\maketitle

\begin{abstract}
Accurately predicting the Pavement Condition Index (PCI), a measure of roadway conditions, from pavement images is crucial for infrastructure maintenance. This study proposes an enhanced version of the Residual Network (ResNet50) architecture, integrated with a Convolutional Block Attention Module (CBAM), to predict PCI directly from pavement images without additional annotations. By incorporating CBAM, the model autonomously prioritizes critical features within the images, improving prediction accuracy. Compared to the original baseline ResNet50 and DenseNet161 architectures, the enhanced ResNet50-CBAM model achieved a significantly lower mean absolute percentage error (MAPE) of 58.16\%, compared to the baseline models that achieved 70.76\% and 65.48\% respectively. These results highlight the potential of using attention mechanisms to refine feature extraction, ultimately enabling more accurate and efficient assessments of pavement conditions. This study emphasizes the importance of targeted feature refinement in advancing automated pavement analysis through attention mechanisms.
\end{abstract}


%
\IEEEpeerreviewmaketitle


\section{Introduction}
\IEEEPARstart{P}{} avement management systems (PMS) are essential for maintaining roadway infrastructure, ensuring safety, and optimizing resource allocation throughout the pavement lifecycle.  These systems rely on accurate assessments of pavement conditions to identify areas requiring repair or rehabilitation and to allocate maintenance resources efficiently. A critical component of this process is the Pavement Condition Index (PCI), which serves as a standardized metric for evaluating pavement health based on observed distresses, severity, and extent \cite{PerezAcebo2018Research,SHOLEVAR2022104190,owor2024pavesam,owor2023image2pci}. Accurate PCI evaluation enables transportation agencies to prioritize maintenance, prevent costly reconstruction, and extend pavement life \cite{Sabouri2025Hybrid}.

Traditionally, PCI has been determined through manual inspections or image-based surveys, where experts visually evaluate pavement conditions and assign scores ranging from 0 (worst condition) to 100 (best condition). While effective, these methods are labor-intensive, costly, and prone to subjectivity, inconsistency, and human error \cite{Elhadidy2019Simplified, Hasibuan2019Study, Shahnazari2012Application}. Recent advancements in automation have aimed to address these challenges, leveraging machine learning (ML) and deep learning (DL) to improve the efficiency and reliability of PCI assessment \cite{Guan2023Deep, Shah2013Development}.

Shallow ML approaches, such as support vector machines, decision trees, and regression models, have shown promise in automating PCI evaluation. These methods offer faster and less labor-intensive alternatives but are constrained by their reliance on feature engineering and limited ability to capture nonlinear patterns in pavement data \cite{Majidifard2020Deep, Piryonesi2020Data, Chandra2013Relationship, Yamany2019Performance,behzadian20221st,kyem2024pavecap}. On the other hand, DL models, particularly convolutional neural networks (CNNs), have demonstrated superior performance by learning intricate patterns directly from raw data without manual feature extraction. Studies integrating algorithms like YOLO and U-net into pavement distress detection have highlighted the potential of DL to revolutionize PMS \cite{Ali2023Predicting, Guerrieri2024Asphalt, Naseri2022Hybrid,aboah2023deepsegmenter}. Despite their advantages, existing DL models often depend on large, manually labeled datasets for training, introducing biases, scalability challenges, and significant costs that limit their broader adoption \cite{Adesunkanmi2024, Mei2020Cost, Sirhan2022Implementation}.

Attention mechanisms have emerged as a powerful tool to address these challenges, enabling DL models to focus on the most relevant regions of an input while ignoring noise and irrelevant features. By dynamically prioritizing critical information, attention mechanisms enhance the accuracy and efficiency of models across various tasks. In pavement distress detection, attention mechanisms can help models identify subtle yet significant patterns in pavement images, improving the precision of PCI prediction without the need for extensive manual data labeling \cite{Correia2022Attention, Philip2023ASENN, Song2019Automatic,aaboahvision,aboah2020smartphone}.

Building on these advantages of attention mechanisms, this study leverages attention mechanisms within an enhanced ResNet50 architecture, incorporating Convolutional Block Attention Modules (CBAM) to enable both spatial and channel-wise focus during model training. By eliminating reliance on manual annotations and enabling self-guided learning, the proposed framework aims to improve the scalability, accuracy, and practicality of automated PCI assessment.
The contributions of this study are as follows: 
\begin{enumerate}
   \item Development of a novel image-to-PCI framework incorporating spatial and channel attention mechanisms to enable accurate, real-time PCI prediction.
   \item Evaluation of the proposed model's performance against state-of-the-art techniques, demonstrating its efficiency and accuracy.
   \item Establishment of a scalable and automated solution that reduces reliance on manual data labeling, addressing key limitations in existing approaches. 
\end{enumerate}

The subsequent sections of the paper are structured as follows: a review of relevant literature sources will be provided, followed by a description of the methodology designed to fulfill the study's objectives. Section 4 will delve into the data and metrics employed in the investigation, while Section 5 will discuss the findings. Subsequently, the conclusion and areas of future research will be outlined.

\section{Related Literature Works}
\subsection{Evolution of PCI Assessment Methods}

The Pavement Condition Index (PCI) assessment process has undergone significant advancements, transitioning from traditional manual methods to automated approaches. Initially, manual visual surveys conducted by field experts or data collection vehicles such as Automated Road Analyzer (ARAN) systems were widely employed \cite{Karim2016Road, Majidifard2020Deep,aboah2019investigation,aboah2023real,kyem2024weather}. While effective, these methods are labor-intensive, costly, and prone to subjectivity, which limits their scalability and efficiency. For instance, the Ohio Department of Transportation reported substantial costs associated with operating ARAN vehicles \cite{Majidifard2020Deep,wang2024gazesam,kwakye2024all}. 
Table 1 provides a summary of evolution in PCI assessment methods.

\begin{table*}[ht]
\centering
\def\arraystretch{1.4}
\setlength{\tabcolsep}{4pt}
\caption{\textbf Summary of Advancements in PCI Assessment Methods}
\begin{tabular}{p{3.5cm}p{3.0cm}p{3.5cm}p{4.5cm}}
\hline
\textbf{Reference} & \textbf{Approach Used} & \textbf{Advancements} & \textbf{Limitations} \\
\hline

(Majidifard et al., 2020; Karim et al., 2016) & 
Manual Visual Surveys, ARAN & 
Initial methods for PCI assessment, operationally effective & 
Labor-intensive, costly, prone to subjectivity, inconsistent, and potentially unsafe for surveyors in traffic \\
\hline

(Chin et al., 2019) & 
Smart Sensor Systems & 
Sustainable and reliable alternative, improved data collection & 
Higher initial implementation costs compared to manual methods, limited automation relative to advanced ML techniques \\
\hline

(Sabouri \& Mohammadi, 2023; Guerrieri et al., 2024) & 
Automated Solutions & 
Reduced reliance on human labor, enhanced efficiency, and reduced subjectivity & 
Limited scalability without advanced computational approaches \\
\hline
\end{tabular}
\end{table*}

\subsection{Shallow Machine Learning Approaches}
Shallow machine learning (ML) methods have played a pivotal role in automating PCI assessments by providing faster and less resource-intensive alternatives to traditional techniques. Various algorithms, including support vector machines (SVM), decision trees, and regression models, have demonstrated moderate success in tasks such as crack detection and PCI prediction. For example, \cite{Li2009Automation,duah2024divneds,denteh2025integrating} developed a crack classification algorithm using SVM, which outperformed back-propagation neural networks (BPNN). \cite{Fujita2017QCAV} and\cite{Wang2017Cracking,kyem2025context,asamoahsaam} employed linear SVM and rectangular SVM models, respectively, to improve pavement crack detection and classification accuracy. Table 2 provides a summary of the advancements and limitations for shallow machine learning approaches for PCI assessment.

Despite these advancements, shallow ML methods rely heavily on manual feature engineering, requiring significant domain expertise to extract relevant features from raw data \cite{Guan2023Deep, Shahnazari2012Application}. This limitation hampers their ability to model complex, nonlinear relationships inherent in pavement datasets. Ensemble approaches and hybrid techniques, such as those proposed by \cite{Naseri2022Hybrid} and \cite{Adesunkanmi2024}, have sought to mitigate these challenges, but shallow ML methods remain constrained by reduced generalization capabilities and scalability issues \cite{Behnood2020Machine, Sabouri2023Hybrid}.

\begin{table*}[ht]
\centering
\def\arraystretch{1.4}
\setlength{\tabcolsep}{4pt}
\caption{\textbf Shallow Machine Learning Approaches for PCI Assessment}
\begin{tabular}{p{3.0cm}p{3.5cm}p{3.5cm}p{4.5cm}}
\hline
\textbf{Reference} & \textbf{Approach Used} & \textbf{Advancements} & \textbf{Limitations} \\
\hline

(Fujita et al., 2017) & 
Support Vector Machine (SVM) & 
Outperformed BPNN for crack classification & 
Requires manual feature engineering; limited ability to model nonlinear relationships \\
\hline

(Shahnazari et al., 2012) & 
Artificial Neural Network (ANN) & 
Superior accuracy compared to traditional models & 
Inefficient at modeling complex non-linear relationships in diverse datasets \\
\hline

(Sabouri \& Mohammad, 2023; Benhood \& Daneshvar, 2020) & 
Linear and Rectangular SVM & 
Improved crack detection and classification accuracy & 
Reduced scalability and generalization; relies heavily on domain-specific feature extraction \\
\hline

(Naseri et al., 2023; Adesunkanmi et al., 2023) & Hybrid Approaches (e.g., Random Forest, Wrapper Selection) & Enhanced accuracy and feature selection methods & 
Requires preprocessing and manual tuning; limited adaptability to diverse pavement conditions \\
\hline
\end{tabular}
\end{table*}

\subsection{Advances in Deep Learning for PCI Estimation}
Deep learning (DL) represents a transformative advancement in pavement condition assessment, offering data-driven, end-to-end learning solutions that eliminate the need for manual feature engineering. Convolutional neural networks (CNNs), in particular, have been widely adopted for their ability to learn intricate patterns directly from raw pavement images \cite{Guan2023Deep, SHOLEVAR2022104190, Piryonesi2020Data, Maniat2021Deep}. \cite{Majidifard2020Deep} integrated U-net and YOLO algorithms to automate crack detection and classification, significantly reducing reliance on human input. Similarly, \cite{Ren2023YOLOv5sM} improved YOLOv5 for pavement distress detection by incorporating cross-layer and cross-scale feature fusion techniques.

Other notable contributions include \cite{Mei2020Cost} ConnCrack algorithm, which utilized generative adversarial networks for pixel-level crack detection, demonstrating superior performance compared to traditional methods like ResNet and VGG-FCN. \cite{Lei2020Automated} further advanced real-time PCI assessment by integrating YOLOv3 with geographic data and decision trees for crack localization and temporal deterioration evaluation. These studies highlight the scalability, efficiency, and accuracy of DL methods in addressing the challenges of traditional and shallow ML approaches. Table 3 presents the summary of findings for deep learning approaches for PCI estimation.

\begin{table*}[ht]
\centering
\def\arraystretch{1.4}
\setlength{\tabcolsep}{4pt}
\caption{\textbf Deep Learning Approaches for PCI Estimation}
\begin{tabular}{p{3.0cm}p{3.5cm}p{3.5cm}p{4.5cm}}
\hline
\textbf{Reference} & \textbf{Approach Used} & \textbf{Advancements} & \textbf{Limitations} \\
\hline

(Piryonesi et al., 2020; Majidifard et al., 2020) & 
U-Net and YOLO Algorithms & 
Automated crack detection, classification, and quantification & Relies heavily on manual annotations; scalability remains a challenge \\
\hline

(Mei \& Gül, 2020; Lei et al., 2020) & Generative Adversarial Networks (GAN) & Pixel-level crack detection with superior accuracy & 
Requires curated datasets: subjectivity in annotations may introduce biases \\
\hline

(Maniat \& Camp, 2021) & Convolutional Neural Networks & Effective crack detection using GSV images & Limited scalability due to annotation dependence and dataset-specific training \\
\hline
\end{tabular}
\end{table*}

\subsection{Addressing Limitations with Attention Mechanisms}
Despite their advantages, existing DL models heavily depend on large, manually annotated datasets for training, which introduces biases, scalability issues, and significant costs \cite{Guerrieri2024Asphalt, Majidifard2020Deep, Mei2020Cost}. Attention mechanisms have emerged as a promising solution to these challenges, enabling DL models to dynamically focus on the most critical features in an input while ignoring irrelevant information. By enhancing learning efficiency and robustness, attention mechanisms significantly improve the performance of DL models in complex tasks such as pavement distress detection \cite{Maniat2021Deep, Correia2022Attention}.

\cite{Correia2022Attention} emphasized the scalability and adaptability of attention models, which allow networks to allocate computational resources to salient features in pavement images. This capability reduces reliance on human annotations and mitigates biases associated with manual data labeling. Initial studies, such as those by \cite{Song2019Automatic} and \cite{Philip2023ASENN}, have demonstrated the potential of attention mechanisms in improving model accuracy and generalization. However, these efforts still relied on partially annotated datasets, underscoring the need for fully self-guided learning frameworks. Table 4 presents the summary of advancements and limitations of attention mechanisms in DL models.

\begin{table*}[ht]
\centering
\def\arraystretch{1.4}
\setlength{\tabcolsep}{4pt}
\caption{\textbf Integration of Attention Mechanisms in DL Models}
\begin{tabular}{p{3.0cm}p{3.5cm}p{3.5cm}p{4.5cm}}
\hline
\textbf{Reference} & \textbf{Approach Used} & \textbf{Advancements} & \textbf{Limitations} \\
\hline

(Correia \& Colombini 2022) & 
Channel and Spatial Attention & 
Enhanced focus on critical features, improved scalability & Still relies partially on annotated datasets for training \\
\hline

(Song et al., 2019; Philip et al., 2023) & Multiscale Feature Attention & Improved accuracy and feature prioritization & Requires additional computational resources; manual data representation hinders generalization \\
\hline
\end{tabular}
\end{table*}

\subsection{The Gap in Literature and Proposed Solution}
Based on the review discussed in this section, there is an apparent need to improve the assessment of pavement conditions. The traditional means of visually inspecting pavement images to evaluate pavement conditions is time-consuming, impracticable, subjective, and inefficient in providing timely rehabilitation and maintenance. The introduction of shallow machine learning made significant improvements toward automation in pavement condition assessment; however, it came with its challenges, such as being unable to learn the complexities of non-linear relationships between pavement condition data and PCI. The advent of DL algorithms increased the efficiency of evaluating pavement conditions but is limited to their reliance on human annotators, which introduces bias in the learning process, thus rendering their results limited in generalizability. All the approaches explored by researchers have been time-consuming, prone to errors, impractical for implementation, and require computational resources. Our study aims to alleviate the need for human annotators, thus reducing the bias in the pavement detection process by building a novel enhanced ResNet model incorporating the CBAM. 

Our model's design was carefully developed to enable the innate differentiation of pavement damages from their surroundings using an attention mechanism. To our knowledge, this is the first effort to associate pavement imagery directly with the PCI without relying on human-labeled data. The performance of our model is compared to other state-of-the-art approaches.  

\section{Methodology}
This paper proposes an improved ResNet50 architecture integrating the CBAM to help the model explicitly learn from pavement images to predict PCI without any auxiliary annotations. This section discusses our baseline models and our proposed model architecture.

\subsection{ResNet50}   
ResNet50, as shown in Fig. 1, is a variant of the Residual Network architecture, which introduces the concept of "skip connections," allowing the network to bypass one or more layers. The fundamental unit of ResNet50 is the residual block, which addresses the vanishing gradient problem associated with DL models, allowing the training of much deeper networks \cite{He2016Deep}. The residual block can be represented mathematically as $F(x) + x$, where x is the input feature map to the residual block, and $F(x)$ denotes the output of the residual block's operations.

\begin{figure}[h]
    \centering
    \includegraphics[width=1\linewidth]{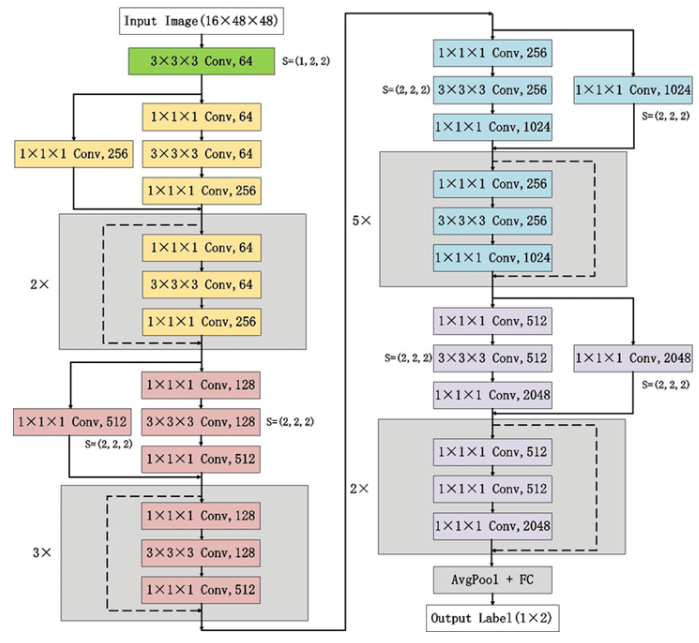}
    \caption{A diagram of ResNet50 architecture}
    \label{fig:enter-label}
\end{figure}

The ResNet50 architecture shown in Fig. 1 starts with an initial convolutional layer, usually with a 3x3 kernel and 64 filters, followed by a max pooling layer. This is followed by four main convolutional stages, each with a series of bottleneck blocks. The first and second stages contain 3 and 4 bottleneck blocks, respectively. The third stage includes 6 bottleneck blocks, while the fourth and final stage contains 3 bottleneck blocks. Each bottleneck block typically contains three layers: a 1x1 convolution for dimensionality reduction (channel reduction), a 3x3 convolution for spatial processing, and a 1x1 convolution for dimensionality restoration (channel restoration). There is also the use of skip connections that add the input of the bottleneck block to its output (residual connections) to help train deeper networks by allowing gradients to flow through the network. The output from the final stage goes through an average pooling layer followed by a fully connected layer, which outputs the final classification scores for the number of classes in the dataset. The structure may also include Batch Normalization and ReLU activation functions after each convolution, except for the final output of the block where the addition with the skip connection occurs before the final ReLU.

\subsection{DenseNet161}
The DenseNet161 model shown in Fig. 2 operates on the principle of feature reuse, where each layer is connected to every other layer in a feed-forward fashion. For each layer, the feature maps of all preceding layers are used as inputs, and its feature maps are used as inputs into all subsequent layers. This connectivity pattern leads to substantial improvements in efficiency and effectiveness. The output of the {\em ith} layer in DenseNet can be mathematically described as:
\begin{equation}
x_i = H_i([x_0, x_1, ..., x_{i-1}])
\end{equation}
{\em $H_i ([x_0,x_1,\ldots,x_{i-1}])$} denotes the concatenation of the feature maps produced in layers 0, 1, …, i-1, and Hi represents the non-linear transformation (including Batch Normalization, ReLU, and Batch Normalization, ReLU, Convolution) applied at the ith layer.

\begin{figure}
    \centering
    \includegraphics[width=0.38\linewidth]{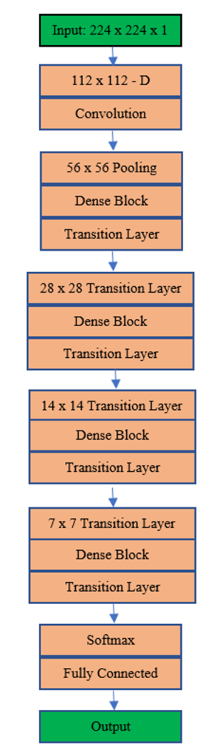}
    \caption{Archtecture for DenseNet161 model}
    \label{fig:enter-label}
\end{figure}

\begin{figure*}[h!]
    \centering
    \includegraphics[width=1\linewidth]{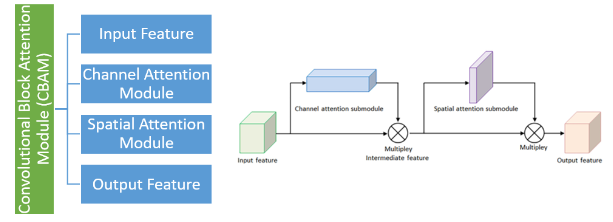}
    \caption{Overview of the Convolutional Block Attention Module (CBAM)}
    \label{fig:enter-label}
\end{figure*}

\subsection{Convolutional Block Attention Module}
The CBAM is an attention mechanism that sequentially infers attention maps along the channel and spatial dimensions. This module can be easily integrated into existing CNN architectures, enhancing their capability to focus on relevant features for a given task. The main idea behind the CBAM is to improve the representation of neural networks by focusing on spatial and channel attention and how important they are in an image. The CBAM (see Figure 3) is designed as a lightweight and flexible module that can be integrated into any CNN architecture without substantial modifications. A CBAM is typically inserted after each convolution block, allowing the network to adaptively adjust its focus on relevant features at different levels of abstraction. This sequential attention process (first channel, then spatial) enables a refined feature refinement strategy, effectively enhancing the model's ability to discriminate between important and unimportant features.

\subsubsection{Channel Attention Module}\label{subsec:channel_attention}
The Channel Attention Module (CAM) identifies the most relevant features across different channels by analyzing the inter-channel relationships and generating a 1D channel attention map. A CAM highlights the ‘what’ aspects of the input images that are most significant for any given classification task by recognizing that each channel acts as a unique feature detector \cite{Zeiler2014Visualizing}. The CAM achieves this by compressing the spatial dimensions of the input feature map and adopting both average-pooling and max-pooling to aggregate spatial information efficiently. The average-pooling method captures the general extent of target objects within the input images. In contrast, the max-pooling provides additional information by emphasizing distinct object features. This dual approach significantly enhances the model's capability to generate more precise channel-wise attention. The mechanism can be formulated as:
\begin{equation}
M_C(F) = \sigma(MLP(\textit{AvgPool}(F)) + MLP(\textit{MaxPool}(F)))
\end{equation}
where $F$ is the input feature map, $M_C(F)$ denotes the channel attention map, $\sigma$ represents the sigmoid function, $\text{AvgPool}$ and $\text{MaxPool}$ are global average pooling and max pooling operations, respectively, and $\text{MLP}$ denotes a multi-layer perceptron with one hidden layer.

The CAM generates two sets of spatial context descriptors from the input feature map through average and max-pooling. These descriptors are processed by a shared MLP with a single hidden layer, where the size of the hidden activation is determined by a reduction ratio to minimize the number of parameters. The MLP outputs are combined using element-wise summation to form the channel attention map. This process is mathematically represented by applying a sigmoid function to the summation of MLP outputs, with the MLP weights being shared between both pooling methods. Including the ReLU activation function after MLP processing further refines the attention mechanism.

\subsubsection{Spatial Attention Module}\label{subsec:spatial_attention}
The Spatial Attention Module (SAM) enhances the focus on specific spatial regions within an image by assessing the inter-spatial relationships of features. This approach contrasts with channel attention by concentrating on the location of informative parts, thereby complementing the channel-focused analysis. SAM identifies these regions by applying average-pooling and max-pooling along the channel axis and then merging the results. This method effectively highlights areas of interest by creating a comprehensive feature descriptor from the pooled information. This feature descriptor is then processed through a convolution layer, which refines and emphasizes the spatial data, resulting in a 2D spatial attention map. This map, processed through a sigmoid function, indicates areas within the image to focus on or suppress, effectively guiding the model's attention spatially. The spatial attention operation can be expressed mathematically as:

\begin{equation}
{\small M_C(F) = \sigma(MLP(AvgPool(F)) + MLP(MaxPool(F)))}
\end{equation}
where $[\cdot;\cdot]$ denotes the concatenation operation, $f^{7 \times 7}$ represents a convolution operation with a $7 \times 7$ filter, and $M_S(F)$ is the spatial attention map.

\subsection{Proposed ResNet50 + CBAM (Ours)}
This paper proposes incorporating the CBAM into a ResNet50 architecture to enable the model to independently differentiate between distress features and their background without any auxiliary annotation guide. This architecture, shown in Figure 4,  combines the depth and residual learning capabilities of ResNet50 with the focused attention mechanisms of the CBAM. Building on the ResNet50 architecture we talked about earlier, the CBAM module (both channel and spatial attention mechanisms) is introduced within the bottleneck to refine the feature maps by focusing on both channel-wise and spatial features. The ResNet50 backbone is structured into four main layers, each with a series of bottleneck modules. These layers progressively increase the channel depth while reducing the spatial dimensions, specifically designed to extract and refine features at different scales.

\begin{figure}
    \centering
    \includegraphics[width=1\linewidth]{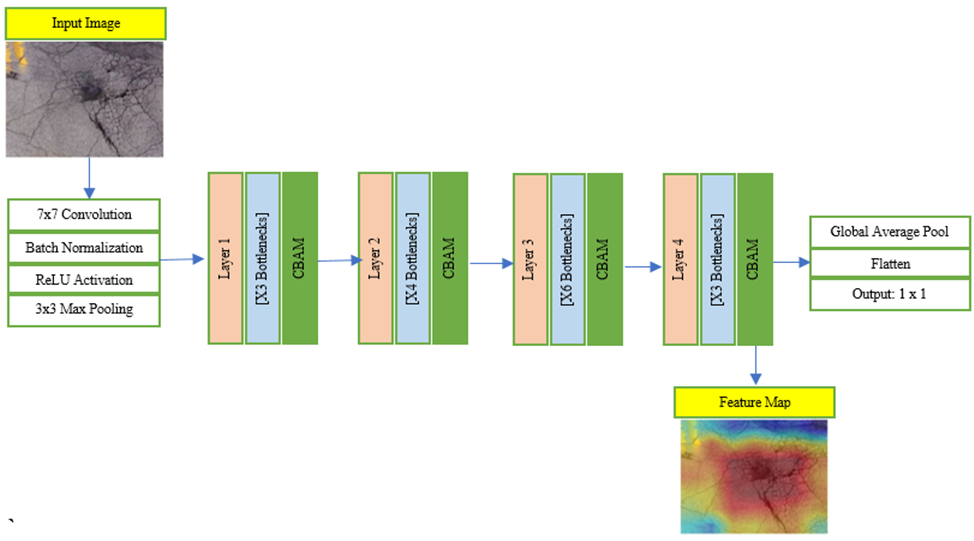}
    \caption{Proposed architecture for ResNet50-CBAM model}
    \label{fig:enter-label}
\end{figure}

The input is first passed through the initial convolutional block during the forward propagation method. It then sequentially goes through Layers 1 to 4, where each layer comprises several bottleneck modules augmented with a CBAM for adaptive feature refinement and distress-specific focus. After the final layer, the feature maps are pooled globally and passed through a fully connected layer to produce the final output. 

\section{Experiments}

\subsection{Dataset}
The dataset used in this study was provided as part of the third DSPS student competition on the application of AI for pavement condition monitoring. The competition tasked participants with employing novel machine learning algorithms to predict the PCI for road sections based on images captured from infrastructure-mounted sensors.  The dataset is publicly available on the DSPS competitions' GitHub page on this link.  The dataset comprises top-down views of pavement captured under varying conditions from three cities: Jefferson City, Missouri; Peoria, and Washington, Illinois. The dataset includes 7704 images labeled with the corresponding PCI, calculated according to ASTM standards, focusing on crack type, extent, and severity. The images were resized to 224x224 pixels, and a series of data augmentation techniques, such as color jittering and random horizontal and vertical flips, were applied to increase the diversity of the training data. The transformations resulted in a total of 30,816 images for our experimentation. These transformations ensure that the models are more robust to variations and ensure generalization, avoiding overfitting. The dataset was split into training and validation sets, with 90\% of the data allocated for training, the remaining 10\% for validation to evaluate the model's performance on unseen data. This is to ensure the generalizability of the trained model. We also used a batch size of 32 for our experimentation. A scheduler was adapted to adjust the learning rate dynamically, optimizing the training process. The model's performance was evaluated on the validation set at the end of each epoch, providing results on the generalization capability and the early stopping to prevent overfitting. This approach ensured that the final model was both accurate and generalizable, meeting the competition's objective of advancing AI-based pavement condition monitoring. This experiment used a V100 GPU and 16 GB of high RAM, utilizing Python 3 and PyTorch framework.

\subsection{Evaluation Metrics}
Evaluation metrics provide a comprehensive understanding of the models’ accuracy and error and a detailed analysis of their prediction capabilities for the pavement condition index. The evaluation metrics used in this study to assess the model’s performance included mean absolute error (MAE), mean absolute percentage error (MAPE), and root mean squared error (RMSE). Lower MAE, MAPE AND RMSE values indicate that the model has a lower spread of errors and is therefore more accurate in its predictions – which highlights how closely aligned the predictions are with the actual data.

\subsection{ Mean Absolute Error (MAE)}
MAE measures the average magnitude of errors in a set of predictions without considering their direction. It provides a straightforward indication of prediction accuracy, with lower values indicating better performance. MAE is defined as:
\begin{equation}
MAE = \frac{1}{n}\sum_{i=1}^n |y_i - \hat{y}_i|
\end{equation}
where $n$ is the number of observations, $y_i$ is the actual value, and $\hat{y}_i$ is the predicted value.

\subsection{Mean Absolute Percentage Error (MAPE)}

MAPE expresses the error as a percentage of the actual values, offering an intuitive understanding of the prediction accuracy in relative terms. It is beneficial for comparing the performance of models across different scales or units. MAPE is calculated as:

\begin{equation}
MAPE = \frac{100\%}{n}\sum_{i=1}^n \left| \frac{y_i-\hat{y}_i}{y_i} \right|
\end{equation}

\subsection{Root Mean Squared Error (RMSE)}
RMSE measures the square root of the average squared differences between the actual and predicted values. It gives a higher weight to larger errors, making it sensitive to outliers. RMSE is a widely used metric for assessing the quality of predictions. It is defined as:
\begin{equation}
RMSE = \sqrt{\frac{1}{n}\sum_{i=1}^n (y_i - \hat{y}_i)^2}
\end{equation}

\section{Results and Discussion}
Our experimentation determined that the ResNet50 + CBAM (our proposed) model outperformed the baseline models—ResNet50 and DenseNet161 models. The results (in Table 3) were calculated by finding the actual and predicted PCI values used to calculate the metrics. These actual and predicted values were determined for the entire validation dataset, and the mean values were calculated for the RMSEs, MAEs, and MAPEs.

\begin{table*}[ht]
    \centering
    \caption{Comparison of results}
    \label{tab:comparison_results}
    \renewcommand{\arraystretch}{1.2}
    \begin{tabular}{lccc}
        \hline
        \textbf{Model} & \textbf{RMSE (\%)} & \textbf{MAE (\%)} & \textbf{MAPE (\%)} \\
        \hline
        ResNet50 & 19.2841 & 14.0296 & 70.7574 \\
        DenseNet161 & 19.1858 & 13.9583 & 65.4751 \\
        \textbf{ResNet50 + CBAM} & \textbf{18.7194} & \textbf{13.9311} & \textbf{58.5616} \\
        \hline
    \end{tabular}
\end{table*}

The comparison among ResNet50, DenseNet161, and ResNet50+CBAM highlights the impact of architectural enhancements and attention mechanisms on model performance. ResNet50+CBAM shows the best performance across all metrics, indicating that the attention mechanism provided by the CBAM effectively improves the model's focus on relevant features, leading to more accurate predictions. DenseNet161 model, which has denser connections, outperforms the base ResNet50 model across all metrics (RMSE, MAE, and MAPE) by a marginal difference. This improvement can be attributed to the architectural distinctions between the two models that impact their learning capabilities and efficiency in feature reuse. On the other hand, the ResNet50 is a lighter weight model, and hence more scalable compared to the DenseNet161 model. As such, due to the negligible differences in their performances, ResNet50 model was chosen for further improvements. The inclusion of CBAM with ResNet50 notably reduces both RMSE and MAPE more significantly than changes in network architecture alone (as seen with DenseNet161), highlighting the impact of targeted feature refinement on prediction accuracy.
The results highlight that the performance improvements brought by the CBAM are achieved with only a slight increase in the number of parameters. This indicates that our proposed model's enhanced ability to focus on relevant features for better predictions is not simply because of adding more parameters (which could naively increase model capacity) but rather due to the strategic refinement of features by the CBAM.

\subsection{Interpretability and Visualisations}
As mentioned, our study employed a ResNet50+CBAM model to analyze pavement images for condition assessment. The input images were processed through the ResNet50+CBAM architecture, resulting in feature maps highlighting key distress indicators, such as cracks and degradation, within the pavement images.

\begin{figure*}
    \centering
    \includegraphics[width=0.9\linewidth]{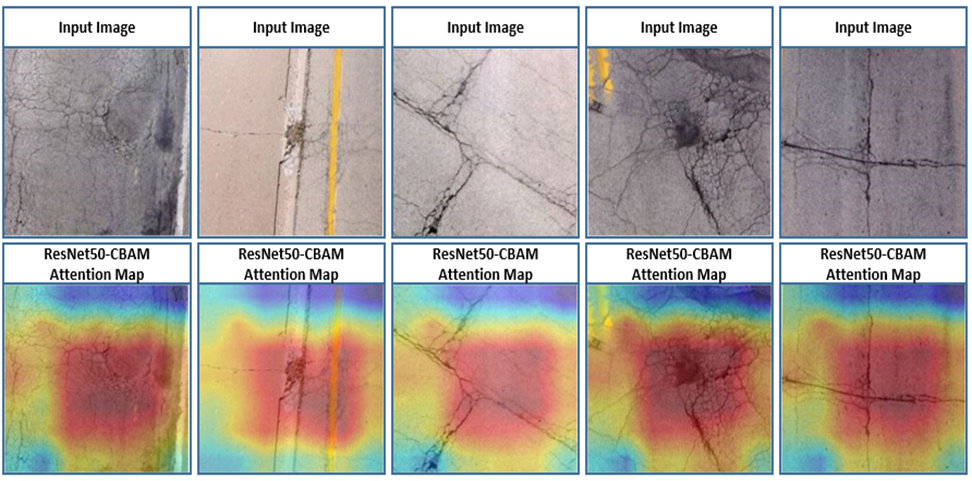}
    \caption{Feature map visualization of results}
    \label{fig:enter-label}
\end{figure*}

\begin{figure*}[h!]
    \centering
    \includegraphics[width=0.8\linewidth]{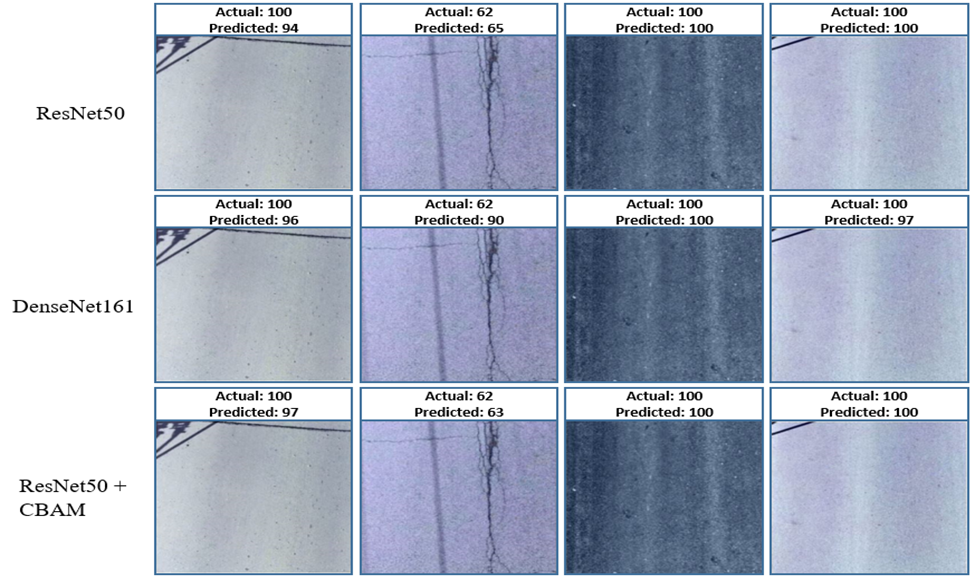}
    \caption{Comparison of best predicted PCI values for ResNet, DenseNet161 \& ResNet50+CBAM models}
    \label{fig:enter-label}
\end{figure*}

Our ResNet50+CBAM model achieved these results without explicit instructions on how these distress indicators should look and by self-learning the weighting features. Color gradients in attention maps typically range from cooler colors like blues and greens to warmer colors like reds and yellows. Cooler colors represent areas of less focus, which are actively suppressed to reduce noise while warmer colors indicate areas of higher focus. These feature maps in Fig. 5, highlighted in red, display the network's attention on the pavement distresses, such as cracks and potholes, showing the region’s most significant for the model's predictions. 

Our proposed architecture with the CBAM focuses on channel and spatial features within the pavement images, enabling the model to prioritize and learn more effectively from relevant features. The channel attention mechanism evaluates the importance of each feature channel, emphasizing the ones most relevant for self-learning. The channel takes the input feature map and performs average and max pooling along the spatial dimensions of the input images. This operation creates two new feature maps that capture how important each channel is based on average and maximum activations across the spatial locations. A shared MLP is applied to the average and max pooled features, which learns to weigh their importance. The results are summed and passed through a sigmoid activation to get a channel attention vector. This vector scales each channel of the original feature map, emphasizing informative channels and suppressing less informative ones – highlighted in yellow, green, and blue in Fig. 5. The spatial attention in our proposed architecture then refines this further by focusing on specific areas within these channels, allowing the model to concentrate on regions of interest like pavement distresses.

Compared to baseline models – ResNet50 and DenseNet161 – our approach improves identifying and concentrating on the target object regions, which correlates with increased accuracy in predicting the PCI. This visual validation (as shown in Fig. 5), alongside quantitative metrics in Table 1, underscores the model's enhanced capability in interpreting and utilizing visual data for pavement analysis. These results indicate that our model did not just memorize the training data but learned to pay attention to the most relevant features crucial for accurate pavement condition assessment. Our proposed model’s ability to focus on relevant regions results from incorporating the CBAM and suggests it contributes to the model’s generalizability. This targeted attention mechanism is a crucial contributor to the improved performance of our model in pavement crack detection and PCI estimation – making it more applicable to unseen pavement conditions.

\begin{figure*}[h!]
    \centering
    \includegraphics[width=0.8\linewidth]{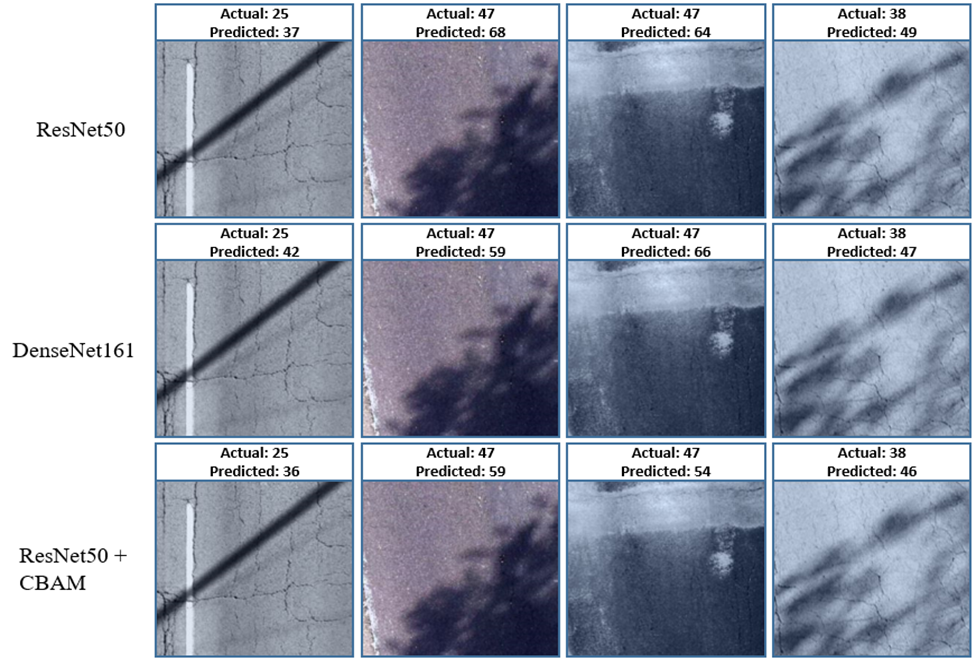}
    \caption{Comparison of worst predicted PCI values for ResNet50, DenseNet161 and ResNet50+CBAM models}
    \label{fig:enter-label}
\end{figure*}

\subsection{Comparative Analysis}

\begin{figure*}
    \centering
    \includegraphics[width=1\linewidth]{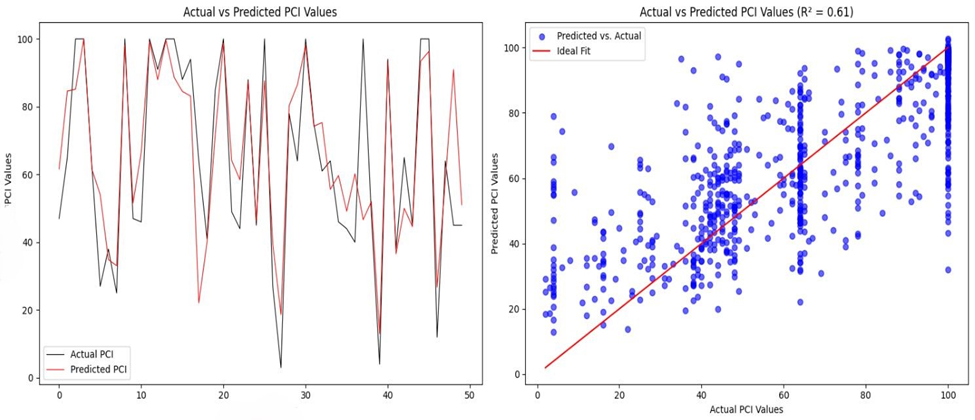}
    \caption{Proposed model performance: (Left) line plot comparing actual and predicted PCI values and (Right) scatter plot showing the correlation between actual and predicted PCI values}
    \label{fig:enter-label}
\end{figure*}

The comparative analysis of the three models — ResNet50, DenseNet161, and ResNet50+CBAM — using their best predictions highlights the impact of attention mechanisms on model performance in this image-based task for a few samples (shown in Fig. 6). 

For the best predictions, ResNet50+CBAM exhibits a finer alignment between the actual and predicted values, indicating that the attention mechanism provides a more detailed interpretation of the images. This improvement is reflected quantitatively in the lowest RMSE, MAE, and particularly in the MAPE scores (Table 1), suggesting greater precision in predictions across varied pavement conditions. This is evidenced in the actual vs. predicted PCI values for the first two images of ResNet50, which are (100 vs. 94) and (62 vs. 65), respectively, as compared to our proposed models, which had (100 vs. 97) and (62 vs. 63), respectively.
In the case of the worst predictions, while all models diverge from the actual values, ResNet50+CBAM maintains closer proximity, again demonstrating the benefits of the CBAM module. A comparison of actual and predicted PCI values in Fig. 7 shows the stack difference in performance between the three models. ResNet50 model shows actual vs. predicted PCI values of (25 vs. 37) and (47 vs. 64) for the first and third sampled images, respectively, compared to our proposed model that shows (25 vs. 36) and (47 vs. 54) for the same sampled images, respectively. This confirms that the CBAM enhances the model's focus even in its worse predictions, likely allowing it to capture more relevant features that contribute to a more accurate prediction. ResNet50+CBAM's ability to self-learn critical details within the pavement images could explain its improved performance metrics compared to ResNet50 and DenseNet161, which lack this targeted attention capability.

To further compare the predicted PCI values against the actual PCI values across different models on the test dataset, we conducted correlation analyses and visualized the results using line plots and scatter plots. Fig. 8 presents the results for the proposed model (Resnet50 + CBAM), Fig. 9 for the ResNet50 baseline model, and Fig. 10 for the DenseNet161 model. In each figure, the line plot (left) compares the actual PCI values (black line) with the predicted PCI values (red line), allowing us to evaluate how closely the models’ predictions align with the ground truth. The scatter plot (right) in each figure depicts the correlation between actual and predicted PCI values, providing an additional measure of model performance. 
Fig. 8 (a) demonstrates that, in most cases, the predicted PCI values (red) follow the general trend of the actual PCI values (black), reflecting our proposed model's ability to predict the PCI values relative to the ground truth. However, some significant deviations occur in a few cases. This suggests that our proposed model may struggle with abrupt transitions or extreme values, indicating limitations in capturing certain patterns or detailed features within the pavement images. The R² value of 0.61, shown in Fig. 8 (b), quantifies the variance in the actual PCI values explained by the model's predictions. While this represents a significant improvement over baseline models, there remains room for enhancing prediction reliability. Further evaluation reveals higher prediction variability at lower PCI scores (i.e., degraded pavements). This variability may stem from the complexity of predicting conditions in severely damaged pavements, where distress features often overlap or blend. It may also reflect potential biases in the training data if low PCI cases are underrepresented or exhibit greater variability in distress types. Furthermore, predictions align better with actual values for higher PCI scores, indicating well-maintained pavements, which often display clearer and more consistent features (e.g., fewer cracks). This confirms that the model is relatively effective at predicting PCI for higher-quality roads but needs further refinement to handle low PCI scores accurately.

Fig. 9 (a) demonstrates that the predicted PCI values (red) generally follow the trend of the actual PCI values (black), indicating the ResNet50 baseline model’s capacity to approximate pavement conditions. However, some deviations highlight the model's limitations in capturing complex features within pavement images. Fig. 9 (b) reports an R² value of 0.57, indicating that 57\% of the variance in actual PCI values is explained by the model's predictions. Although this R² value suggests moderate predictive strength, it is lower than our proposed model, stressing the need for further refinement. Additionally, the analysis reveals higher variability at lower PCI scores, reflecting the baseline model's difficulty in accurately predicting conditions for severely distressed pavements.

\begin{figure*}
    \centering
    \includegraphics[width=1\linewidth]{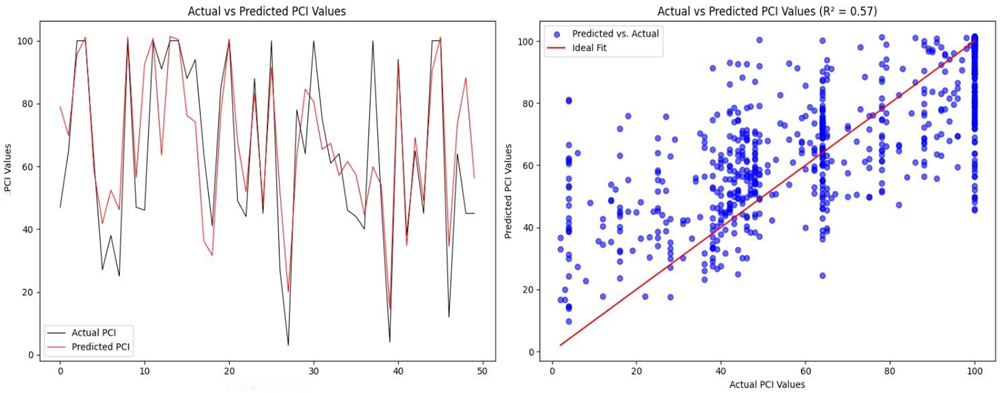}
    \caption{Resnet50 model performance: (Left) line plot comparing actual and predicted PCI values and (Right) scatter plot showing the correlation between actual and predicted PCI values}
    \label{fig:enter-label}
\end{figure*}

The performance of the Densenet161 baseline model is lower compared to our proposed model. Fig. 10 (a) shows the predicted PCI values significantly deviating from the actual values, especially during sharp transitions in many cases. The misalignment may suggest that this model struggles more with capturing complex pavement features. 

\begin{figure*}
    \centering
    \includegraphics[width=1\linewidth]{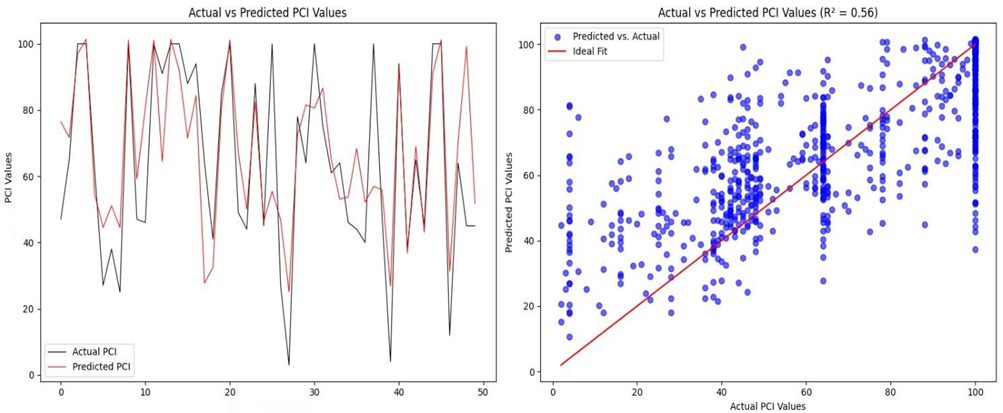}
    \caption{DenseNet161 model performance: (Left) line plot comparing actual and predicted PCI values and (Right) scatter plot showing the correlation between actual and predicted PCI values}
    \label{fig:enter-label}
\end{figure*}

The scatter points in Fig. 10 (b) show greater dispersion, especially where there are lower PCI scores. That, coupled with a lower R² value of 0.56 (compared to our proposed model with an R² value of 0.56) results in reduced prediction reliability for degraded pavements. This model demonstrates weaker alignment with the ideal fit line, generally reflecting less accurate predictions.

\section{Conclusion and Future Work}
In this paper, we presented an enhanced ResNet50 model integrated with the CBAM for directly predicting the PCI from pavement images. Our approach combines the feature extraction capabilities of ResNet50 with the feature refinement power of CBAM's spatial and channel attention mechanisms, enabling the model to focus on critical pavement features while reducing noise. The dataset used for this study was sourced from the DSPS student competition, consisting of 7,704 pavement images with varying distress conditions. We demonstrated that our proposed ResNet50+CBAM model sets a new benchmark in automated PCI prediction. The model achieved a relatively lower mean absolute percentage error (MAPE) of 58.16\% compared to the baseline ResNet50 and DenseNet161 models, which achieved 70.76\% and 65.48\%, respectively.  Our proposed model is reliable for general PCI predictions, especially for well-maintained roads, and represents a significant improvement over baseline methods. However, its variability in predicting low PCI scores and handling edge cases suggests a need for targeted enhancements to improve its utility for critical pavement management tasks.

Despite its strengths, the proposed model faces certain challenges. Prediction accuracy decreases in edge cases, such as heavily degraded pavements or images taken under extreme lighting conditions. Additionally, the computational requirements, including the use of high-end GPUs, may limit the accessibility of this approach for widespread adoption. 

Future work can explore designing physics-informed loss functions. These loss functions would incorporate physical laws governing how pavements degrade over time. This would guide the network during training to learn more realistic and transferable representations of pavement conditions. The model could also become less reliant on specific training data and perform better on unseen pavement conditions, leading to a more robust and adaptable pavement assessment tool. Also, the dataset can be broadened to include more diverse conditions, including edge cases, to improve the model’s generalizability and performance under challenging scenarios.

\section*{Acknowledgments}
This should be a simple paragraph before the References to thank those individuals and institutions who have supported your work on this article.



%





\ifCLASSOPTIONcaptionsoff
  \newpage
\fi





\bibliographystyle{IEEEtran}
\bibliography{IEEEabrv,Bibliography}

\vfill


\end{document}